# A Scalable Framework for logP Prediction: From Terabyte-Scale Data Integration to Interpretable Ensemble Modeling

Malikussaid
School of Computing, Telkom University
Bandung, Indonesia
malikussaid@student.telkomuniversity.ac.id

Septian Caesar Floresko
School of Computing, Telkom University
Bandung, Indonesia
septiancaesar@student.telkomuniversity.ac.id

Ade Romadhony
School of Computing, Telkom University
Bandung, Indonesia
aderomadhony@telkomuniversity.ac.id

Isman Kurniawan
School of Computing, Telkom University
Bandung, Indonesia
ismankrn@telkomuniversity.ac.id

Warih Maharani
School of Computing, Telkom University
Bandung, Indonesia
wmaharani@telkomuniversity.ac.id

Hilal Hudan Nuha
School of Computing, Telkom University
Bandung, Indonesia
hilalnuha@telkomuniversity.ac.id

*Abstract*—The accurate prediction of molecular lipophilicity logP remains a cornerstone challenge in computational drug discovery, influencing multiple absorption, distribution, metabolism, excretion, and toxicity parameters. This study presents a large-scale predictive modeling framework for logP prediction using 426850 bioactive compounds rigorously curated from the intersection of three authoritative chemical databases: PubChem, ChEMBL, and eMolecules. We developed a novel computational infrastructure to address the data integration challenge, reducing processing time from a projected over 100 days to 3.2 hours through byte-offset indexing architecture, a 740-fold improvement. Our comprehensive analysis revealed critical insights into the multivariate nature of lipophilicity: while molecular weight exhibited weak bivariate correlation with logP, SHAP analysis on ensemble models identified it as the single most important predictor globally. We systematically evaluated multiple modeling approaches, discovering that linear models suffered from inherent heteroskedasticity that classical remediation strategies, including weighted least squares and Box-Cox transformation, failed to address. Tree-based ensemble methods, including Random Forest and XGBoost, proved inherently robust to this violation, achieving an R-squared of 0.765 and RMSE of 0.731 logP units on the test set. Furthermore, a stratified modeling strategy, employing specialized models for drug-like molecules (91 percent of dataset) and extreme cases (nine percent), achieved optimal performance: an RMSE of 0.838 for the drug-like subset and an R-squared of 0.767 for extreme molecules, the highest of all evaluated approaches. These findings provide actionable guidance for molecular design, establish robust baselines for lipophilicity prediction using only 2D descriptors, and demonstrate that well-curated, descriptor-based ensemble models remain competitive with state-of-the-art graph neural network architectures.

## I. INTRODUCTION

### A. The Persistent Challenge of Property Prediction in Drug Development

The discovery and development of new pharmaceutical agents remains one of the most resource-intensive endeavors in modern science, with timelines frequently exceeding a decade and costs surpassing billions of dollars per approved therapeutic compound [1]. At the heart of this challenge lies the accurate prediction of molecular properties that govern a drug candidate's behavior in biological systems. Among these properties, lipophilicity—quantified as the octanol-water partition coefficient (logP)—holds a position of fundamental importance. This single physicochemical parameter serves as a determinant of multiple critical absorption, distribution, metabolism, excretion, and toxicity (ADMET) endpoints, including membrane permeability, tissue distribution, metabolic stability, and elimination pathways [1].

The centrality of lipophilicity in medicinal chemistry is perhaps best exemplified by its inclusion in Lipinski's seminal "Rule of Five," which established logP ≤ 5 as one of four criteria for predicting oral bioavailability [2]. However, the role of logP has evolved substantially beyond this simple threshold filter. In the era of artificial intelligence-driven drug discovery, lipophilicity has become an active optimization parameter, with contemporary medicinal chemists seeking to balance multiple physicochemical properties simultaneously to enhance overall therapeutic quality and clinical success rates [1]. This optimization mandate extends to next-generation therapeutic modalities, where careful tuning of lipophilicity is recognized as essential for designing advanced pharmaceuticals, including proteolysis-targeting chimeras (PROTACs) and other complex molecular architectures.

The practical value of accurate computational logP prediction is evident: it enables high-throughput virtual screening of vast chemical libraries, prioritizing synthesis and experimental testing resources toward the most promising candidates. However, despite decades of research and the development of numerous empirical, fragment-based, and machine learning approaches, logP prediction remains challenging. Experimental measurements themselves exhibit reproducibility limits of approximately ±0.3 logP units [3], while computational methods must contend with the enormous diversity of chemical space and the complex, multivariate determinants of octanol-water partitioning behavior.

### B. The Data Quality Imperative in Cheminformatics

The accelerating integration of machine learning methodologies into molecular property prediction has created unprecedented opportunities and equally significant challenges. Public chemical databases—most notably ChEMBL [4], [5], [6], [7] and PubChem [8], [9], [10]—have become indispensable resources, providing access to millions of bioactive molecular structures and associated experimental data. These repositories aggregate heterogeneous information from diverse sources, including high-throughput screening

The work reported here is supported by Telkom University under Basic and Applied Research Grant No. 056/LIT06/PPM-LIT/2025. Any opinions, findings, and conclusions or recommendations expressed in this paper are those of the authors and do not necessarily reflect the view of Telkom University.

campaigns, patent literature, medicinal chemistry publications, and commercial chemical suppliers.

However, this aggregation process introduces substantial data quality concerns that directly impact the reliability of computational models. Recent investigations have documented error rates as high as 8% for chemical structures in certain medicinal chemistry databases, with issues ranging from incorrect stereochemistry to duplicate entries and inconsistent property annotations. The growing recognition of these challenges has prompted explicit calls from domain experts, including researchers at the U.S. Environmental Protection Agency, for "automated and manual quality curation procedures" to serve as essential pillars for the future of cheminformatics. A 2025 perspective article in the Journal of Cheminformatics stated unambiguously that "poor data quality is a major concern with the rise of machine-learning and AI methods that aggregate these error-prone Internet resources, potentially propagating errors into computational models such as QSARs" [11].

These concerns are not merely academic. For machine learning models, which learn patterns directly from training data, systematic errors or biases in the input dataset will be faithfully reproduced—and potentially amplified—in model predictions. The task of large-scale data integration and curation is therefore not routine preprocessing but a critical research challenge that fundamentally determines the validity of any downstream predictive model.

*C. Research Objectives and Hypotheses*

This study was conceived to address both the predictive modeling challenge and the data integration imperative through a comprehensive, end-to-end computational workflow. We established three primary research questions:

- **RQ1:** What is the statistical distribution of key physicochemical properties (molecular weight, logP, topological polar surface area) in the intersection of multiple authoritative bioactive chemical databases?
- **RQ2:** What correlations exist between these properties, and to what extent does multicollinearity affect feature-target relationships and model interpretation?
- **RQ3:** Can predictive models for logP achieve $R^2 > 0.6$ using computationally inexpensive two-dimensional molecular descriptors derived from multi-source integrated data?

Corresponding to these research questions, we formulated three testable hypotheses:

- **H1:** Molecular weight (MolWt) exhibits a statistically significant positive correlation with logP, as larger molecules tend to possess greater hydrophobic surface area.
- **H2:** Topological polar surface area (TPSA) is negatively correlated with logP, reflecting the inverse relationship between polar character and lipophilicity, with implications for blood-brain barrier penetration and oral bioavailability.
- **H3:** A multivariate regression model employing a combination of structural molecular descriptors can predict logP with accuracy exceeding $R^2 = 0.6$, a threshold consistent with published quantitative structure-activity relationship (QSAR) models for physicochemical property prediction.

The original project scope anticipated working with approximately 1,000 molecules from a single database source. However, to ensure the generalizability and scientific rigor necessary for publication-quality research, we expanded this scope dramatically, integrating data from three complementary sources and curating a final dataset of 426,850 validated bioactive compounds. This expansion, while introducing substantial computational and methodological challenges, was essential for developing models robust to diverse chemical scaffolds and representative of pharmaceutical-relevant chemical space.

*D. Contributions and Significance*

This work delivers four principal contributions to the field of computational medicinal chemistry:

First, we present a scalable computational infrastructure for multi-terabyte chemical database integration that transforms an intractable $O(N \times M)$ complexity problem into a practical $O(N + M)$ solution through byte-offset indexing. This architectural innovation enabled the processing of 176 million PubChem compound records in hours rather than months.

Second, we provide a rigorous statistical analysis demonstrating that heteroskedasticity—non-constant error variance across the logP range—is an inherent limitation of linear regression models for lipophilicity prediction. We show that classical remediation strategies fail to address this violation and that tree-based ensemble methods offer a principled, robust alternative.

Third, through SHAP (SHapley Additive exPlanations) analysis [12], we resolve apparent paradoxes in feature-target relationships, demonstrating that molecular weight, despite exhibiting weak bivariate correlation with logP, is the single most important predictor when multivariate suppression effects are properly accounted for. This finding has direct implications for molecular design strategy in lead optimization.

Fourth, we demonstrate that a stratified modeling approach—employing specialized models for drug-like molecules versus extreme cases—achieves optimal performance across the full chemical space, with the extreme-molecule model attaining the highest $R^2$ value (0.767) of any approach evaluated.

The remainder of this paper is organized as follows: Section II describes our comprehensive methodology, from multi-source data acquisition through stratified model development. Section III presents detailed results, including dataset characterization, model performance metrics, and feature importance analysis. Section IV discusses the interpretation and significance of our findings, contextualizes performance relative to state-of-the-art approaches, and addresses limitations. Section V concludes with actionable recommendations for both molecular design practitioners and computational modelers, alongside directions for future research.

II. METHODOLOGY

*A. Research Design and Workflow Architecture*

This investigation employed a systematic, multi-phase workflow designed to address both the data integration challenge and the predictive modeling objective. Figure 1 presents the complete analytical pipeline, which progressed through six major stages: (1) multi-source data acquisition, (2) molecular identifier extraction and intersection analysis, (3) large-scale data extraction using novel indexing infrastructure, (4) feature engineering and dataset construction, (5) exploratory data analysis and hypothesis testing, and (6) comparative modeling with statistical validation.

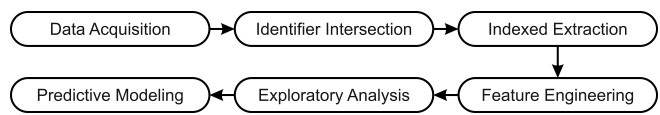

Fig. 1. The 6-phase computational workflow

The workflow was implemented primarily in Python 3.11, leveraging the scientific computing ecosystem including NumPy 2.1.3 and Pandas 2.3.3 for numerical operations and data manipulation, RDKit 2024.03 [16] for cheminformatics operations, and scikit-learn 1.5.0 for machine learning implementations. All analyses were performed on desktop workstations equipped with Intel Core i7 processors, 32 GB RAM, and solid-state storage, demonstrating that large-scale cheminformatics is feasible without specialized high-performance computing infrastructure when appropriate algorithmic strategies are employed.

*B. Multi-Source Data Acquisition and Integration*

*1) Database Selection and Rationale*

We selected three complementary chemical repositories based on orthogonal strengths: comprehensive property coverage (PubChem), rigorous bioactivity curation (ChEMBL), and commercial availability validation (eMolecules).

- *PubChem* [8], [9], [10], maintained by the National Center for Biotechnology Information (NCBI), represents the world's largest freely accessible repository of chemical structures and their properties. As of October 2024, the PubChem Compound collection contained over 176 million unique chemical entities distributed across 354 Structure Data Format (SDF) files totaling multiple terabytes of uncompressed data. PubChem's value for our study derives from its extensive pre-computed molecular descriptors, including XLOGP3 [3]—an empirically validated logP calculation method that serves as our target variable—and comprehensive physicochemical annotations for each compound.
- *ChEMBL* [4], [5], [6], [7], developed and curated by the European Molecular Biology Laboratory's European Bioinformatics Institute (EMBL-EBI), focuses explicitly on bioactive molecules with documented activity against biological targets. ChEMBL's manual curation process and strict quality control standards ensure high-confidence chemical structures, making it an ideal source for validating the biological relevance of molecules in our dataset. The intersection with ChEMBL guarantees that molecules in our final dataset have demonstrated bioactivity, not merely computational existence.
- *eMolecules* [13], a commercial chemical supplier database, provides a third dimension of validation: synthetic accessibility. Molecules present in eMolecules are commercially available from chemical vendors, confirming that they are not merely theoretical constructs but can be procured for experimental validation. This practical constraint ensures our model's applicability to real-world drug discovery workflows.

The strategic rationale for three-source integration is triangulation-based quality assurance. A molecule appearing in all three databases has been (1) comprehensively annotated in a public repository (PubChem), (2) validated as bioactive through experimental screening (ChEMBL), and (3) confirmed as synthetically accessible (eMolecules). This multi-source verification substantially reduces the risk of including erroneous structures, computationally spurious entries, or inaccessible compounds that would compromise model validity.

*2) Automated Acquisition Infrastructure*

The PubChem Compound database is distributed via an FTP server as 354 compressed SDF archives (.sdf.gz format), with individual files ranging from hundreds of megabytes to over 3 GB when compressed. Manual download was impractical, necessitating development of PubChem_Downloader.py, a Python automation script implementing several defensive programming strategies:

- Parallel downloads using aria2c, a high-performance download manager supporting resumable transfers and multi-connection downloads
- MD5 checksum verification for every downloaded file, ensuring bit-perfect data integrity by comparing locally calculated hash values against server-provided checksums
- Automatic retry logic with exponential backoff to handle network interruptions gracefully
- Progress tracking using the tqdm library for real-time monitoring of download status and estimated completion time

This infrastructure proved essential during the three-week download period, as network outages occurred multiple times. The resumable download capability prevented data loss and eliminated redundant transfers that would have wasted server bandwidth. Each successfully downloaded file was verified against its MD5 checksum; corrupted files were automatically detected and re-queued for download.

ChEMBL and eMolecules datasets, being substantially smaller (10-20 GB each), were acquired using conventional HTTP methods. All three databases provided data in SDF format, ensuring structural consistency and enabling unified processing pipelines.

*3) The Data Integration Crisis: Computational Complexity*

The initial integration strategy appeared straightforward: extract unique molecular identifiers from all three databases, compute their intersection, then retrieve full records for common molecules from PubChem. We initially selected InChIKey—a 27-character hashed representation of the IUPAC International Chemical Identifier (InChI)—as our molecular identifier, as it is specifically designed for database lookups and search indexing applications [14].

We developed Inchikey_compare.py to identify the intersection of InChIKeys across databases. This succeeded rapidly, identifying 477,123 molecules common to both ChEMBL and eMolecules. The computational crisis emerged in the subsequent step: locating these 477,123 target molecules within PubChem's 354 massive SDF files containing a total of 176 million compounds.

Our first implementation, SDF_lookup.py, employed a nested-loop search algorithm: for each target InChIKey, scan sequentially through all 354 source files until the molecule is found. This algorithm has computational complexity $O(N \times M)$, where $N$ represents the number of target molecules (477,123) and $M$ represents the total size of the search space (176 million records distributed across 354 files). Even under optimistic assumptions of 1,000 molecules scanned per second, the runtime projection exceeded 100 days. Pilot tests on the first PubChem file confirmed this estimate: processing 500,000 molecules required over 2 hours. This approach was not merely slow—it was computationally infeasible for completion within any reasonable project timeline.

*4) Architectural Pivot: Byte-Offset Indexing*

Recognition that brute-force search was fundamentally flawed prompted a complete architectural redesign. The key insight was to invert the problem: rather than searching files repeatedly for each target molecule, scan all source files exactly once to build a persistent index mapping each molecule's identifier to its precise byte location within the original files.

We implemented this as SDF_indexer.py, employing a parallel processing architecture:

```
def build_index(sdf_files, num_workers=16):
    with multiprocessing.Pool(num_workers)
        as pool:
        partial_indices =
        pool.map(index_single_file, sdf_files)

    global_index = {}
    for partial_index in partial_indices:
        global_index.update(partial_index)

    return global_index
    # {identifier: (filename, byte_offset)}
```

Each worker process independently indexes one SDF file by reading it line-by-line, recording the byte offset at which each molecule begins (obtained via Python's file tell() method), extracting the molecule's identifier, and storing the mapping. The main process aggregates these partial indices into a single comprehensive index.

The indexing process consumed approximately 12 hours on our hardware (Intel Core i7-8700K, 6 cores/12 threads, HDD storage), producing an 11 GB CSV file containing 176,929,690 entries mapping InChIKeys to their exact file locations. Critically, this is a one-time computational investment. Once constructed, the index enables molecule retrieval through simple file seek operations—an $O(1)$ operation per target molecule rather than $O(M)$ sequential scanning.

The new extraction script, Inchi_extractor.py, leverages this index through a streamlined workflow: load the entire index into memory as a Python dictionary (consuming approximately 28 GB RAM), group target molecules by their source files to optimize disk I/O, and for each file containing targets, perform sorted seeks to minimize disk head movement. The extraction of 477,123 molecules completed in 3.2 hours, compared to the projected 100+ days for brute-force search. This represents a 740-fold speedup achieved purely through algorithmic redesign, demonstrating the critical importance of computational complexity analysis in data-intensive science.

*5) The InChIKey Collision Crisis and Pivot to Full InChI*

During validation of extracted molecules, we discovered a fundamental limitation that necessitated complete pipeline reconstruction. InChIKey, despite being an IUPAC-standardized identifier, is fundamentally a hash function—a 27-character representation derived from the potentially hundreds-of-characters-long full InChI string through SHA-256 hashing [14]. Hash functions inherently permit collisions: different inputs producing identical outputs.

While InChIKey collisions are rare (estimated probability of approximately $10^{-15}$ for randomly generated molecules [14]), when operating at the scale of 176 million real-world chemical structures, rare events become statistical certainties. During our validation process—which cross-checked InChIKeys against full InChI strings present in PubChem data fields—we detected multiple instances where structurally distinct molecules (confirmed by differing 2D connectivity diagrams and full InChI representations) shared identical InChIKeys.

For a predictive modeling study, such collisions are scientifically unacceptable. A single InChIKey incorrectly mapping to multiple structures with potentially different logP values introduces label noise that is impossible to detect or correct during model training. The canonical example is the Spongistatin I stereoisomer pair: two distinct molecular configurations possessing different full InChI strings but resolving to the identical InChIKey. For properties sensitive to stereochemistry, such collisions corrupt ground truth labels.

The solution required abandoning InChIKey entirely and rebuilding the complete pipeline using full InChI strings—the comprehensive, collision-free canonical representation defined by IUPAC [14]. Full InChI strings are verbose (typically 100-500 characters) and unsuited for human reading, but they guarantee absolute molecular uniqueness: identical InChI strings correspond to identical chemical entities by definition.

This decision cascaded through every pipeline component:

- SDF_generateInchikey_v2.py was replaced with SDF_generateInchi.py, modified to extract full InChI strings using RDKit's Chem.MolToInchi() function [16]
- Inchikey_compare.py became Inchi_compare.py, operating on full string comparisons
- SDF_indexer.py was updated to use full InChI as dictionary keys, increasing index size from 11 GB to 14 GB
- Inchi_extractor.py incorporated full InChI verification during extraction

The intersection recomputation using full InChI strings yielded 435,413 molecules common to all three databases. This became our verified integration set. While the rework required approximately one week of calendar time, it was non-negotiable from a scientific integrity standpoint: every molecule in the final dataset possesses a unique, IUPAC-validated identifier with zero risk of structural ambiguity.

## C. Feature Engineering and Dataset Construction

### 1) Parallel Transformation Pipeline

With the verified common_molecules.sdf file secured (435,413 molecules, approximately 8 GB uncompressed), we faced the challenge of converting semi-structured SDF format into tabular data suitable for machine learning. The SDF format represents each molecule as a multi-line text block containing (1) a connection table specifying atoms, bonds, and coordinates, (2) zero or more data fields enclosed in tagged sections, and (3) a terminator line ($$$$).

We developed SDF_transform.py employing a producer-consumer parallel architecture to maximize throughput:

```
def transform_sdf_to_csv
    (input_sdf, num_workers=16):
    with multiprocessing.Pool(num_workers)
         as pool:
        results = pool.imap_unordered(
                process_mol_block,
                read_sdf_blocks(input_sdf),
                chunksize=1000)

        for mol_data in results:
            if mol_data is not None:
                write_to_csv(mol_data)
```

The producer function (read_sdf_blocks) reads the SDF file line-by-line, accumulating text into blocks delimited by $$$$ markers, and yields these blocks to the worker pool. Each consumer process parses a molecule block using RDKit's ForwardSDMolSupplier (essential for preserving SDF property fields, unlike the simpler MolFromMolBlock which reads only structural information), calculates molecular descriptors, and returns a dictionary of features and target values.

A critical bug in early development illustrates the importance of thoroughly understanding library behavior: initial versions used Chem.MolFromMolBlock(), which reads molecular structure but ignores all SDF property fields. This resulted in complete loss of the target variable (PUBCHEM_XLOGP3), producing an apparently successful but entirely useless dataset with NaN values for logP. The fix required switching to ForwardSDMolSupplier, which explicitly preserves property annotations. This seemingly minor implementation detail was the difference between success and failure.

### 2) Target Variable and Feature Selection

Our target variable, logP_target, was extracted from the PUBCHEM_XLOGP3 field in each SDF record. XLOGP3 is PubChem's implementation of a well-validated empirical logP calculation method [3] trained on experimental octanol-water partition coefficients. Of the 435,413 extracted molecules, 8,563 (1.97%) lacked this field—likely reflecting computational failures for structurally unusual molecules—and were excluded, yielding a final dataset of 426,850 molecules.

For predictor variables, we selected eight two-dimensional molecular descriptors. These eight descriptors were used for the initial exploratory data analysis (Section III.A). However, as detailed in Section II.D, HeavyAtomCount was subsequently removed from all modeling tasks due to extreme collinearity, leaving a final set of seven descriptors for all modeling and interpretation (Section III.B onwards) calculable directly from the molecular graph without three-dimensional conformational analysis:

1. *MolWt* (Molecular Weight, Da): Overall molecular size
2. *TPSA* (Topological Polar Surface Area [15], Ų): Sum of surface areas of polar atoms (O, N), a validated proxy for membrane permeability
3. *NumHDonors*: Count of hydrogen bond donor groups (NH, OH)
4. *NumHAcceptors*: Count of hydrogen bond acceptor sites (O, N)
5. *NumRotatableBonds*: Count of freely rotatable single bonds, measuring conformational flexibility
6. *NumAromaticRings*: Count of aromatic ring systems
7. *FractionCSP3*: Fraction of sp³-hybridized carbons, quantifying molecular "three-dimensionality" versus "flatness"
8. *HeavyAtomCount*: Total non-hydrogen atoms

This restriction to 2D descriptors was methodologically deliberate: such descriptors are computationally inexpensive (milliseconds per molecule versus seconds for 3D conformer generation), making them suitable for virtual screening of libraries containing millions of compounds. All descriptors were computed using RDKit's Descriptors module [16], ensuring consistency and reproducibility. The final dataset, Computed_Molecules.csv, contained 426,850 rows and 12 columns: three identifier columns (InChIKey, SMILES, Original_InChI), one target column (logP_target), and eight feature columns, with zero missing values—a testament to rigorous quality control.

## D. Exploratory Data Analysis and Statistical Characterization

### 1) Lipinski's Rule of Five Compliance Analysis

Before modeling, we characterized the pharmaceutical relevance of our dataset using Lipinski's Rule of Five [2], which predicts oral bioavailability through four criteria: molecular weight ≤ 500 Da, logP ≤ 5, hydrogen bond donors ≤ 5, and hydrogen bond acceptors ≤ 10. Analysis revealed that 91.0% of the 426,850 molecules satisfied all four criteria simultaneously, substantially exceeding typical bioactivity database pass rates of 60-70% [15]. Individual criterion compliance was: MolWt ≤ 500 (97.05%), logP ≤ 5 (93.58%), NumHDonors ≤ 5 (99.61%), and NumHAcceptors ≤ 10 (99.40%).

Figure 2 presents the Lipinski compliance analysis, demonstrating the marked enrichment for drug-like chemical space in our integrated dataset.

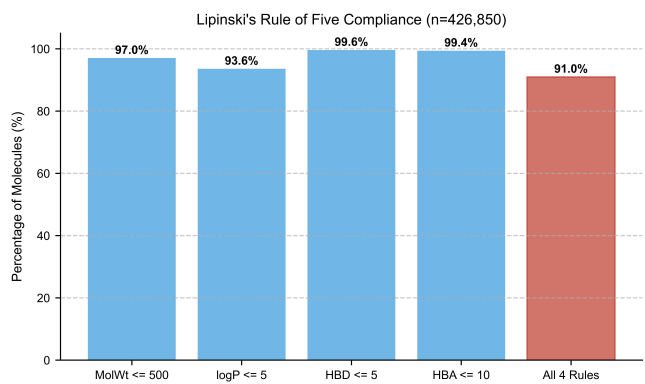

Fig. 2. Overall dataset compliance with Lipinski's Rule of Five

The median molecular weight was 347.3 Da (interquartile range: 295.8-403.5 Da), and median logP was 2.90 (IQR:

2.00-3.80), both centered within optimal ranges for oral drug candidates. The logP distribution exhibited slight negative skewness (-0.775), with a longer tail toward hydrophilic extremes (minimum: -35.3) than lipophilic extremes (maximum: 15.3), characteristic of bioactive compound collections where extreme lipophilicity correlates with off-target toxicity.

*2) Correlation Analysis and Multicollinearity Detection*

Pearson correlation analysis quantified feature-target and feature-feature relationships. Table I summarizes correlations between all eight features and the logP target, ranked by absolute magnitude.

TABLE I. FEATURE-TARGET CORRELATIONS RANKED BY ABSOLUTE MAGNITUDE

| Rank | Feature | Pearson $r$ | $p$-value |
|---|---|---|---|
| 1 | NumAromaticRings | +0.428 | <0.0001 |
| 2 | TPSA | -0.360 | <0.0001 |
| 3 | FractionCSP3 | -0.295 | <0.0001 |
| 4 | NumHDonors | -0.275 | <0.0001 |
| 5 | NumHAcceptors | -0.263 | <0.0001 |
| 6 | MolWt | +0.146 | <0.0001 |
| 7 | HeavyAtomCount | +0.139 | <0.0001 |
| 8 | NumRotatableBonds | -0.050 | <0.0001 |

All correlations achieved statistical significance ($p < 0.0001$) due to the large sample size ($n = 426,850$), but practical significance varied substantially. NumAromaticRings emerged as the strongest single predictor ($r = 0.428$, $R^2 = 0.183$), consistent with chemical intuition that aromatic systems contribute hydrophobic character. TPSA exhibited the expected negative relationship ($r = -0.360$), confirming that increased polar surface area reduces lipophilicity.

Surprisingly, molecular weight showed only weak correlation with logP ($r = +0.146$, $R^2 = 0.021$), seemingly contradicting medicinal chemistry principles that larger molecules should be more lipophilic. This apparent paradox would later be resolved through multivariate analysis (Section III-C).

More concerning was the detection of extreme multicollinearity among features. Figure 3 presents the complete correlation matrix, revealing several problematic feature pairs:

- MolWt vs. HeavyAtomCount: $r = +0.975$ (near-perfect redundancy)
- TPSA vs. NumHAcceptors: $r = +0.797$
- TPSA vs. NumHDonors: $r = +0.776$
- MolWt vs. NumRotatableBonds: $r = +0.705$

Variance Inflation Factor (VIF) analysis [17] quantified multicollinearity severity. VIF values exceeding 10 indicate problematic collinearity; six of eight features exhibited VIF > 10, with MolWt (VIF = 281.7) and HeavyAtomCount (VIF = 362.6) showing extreme inflation. This severe multicollinearity violates assumptions of ordinary least squares regression, inflating standard errors of coefficient estimates and destabilizing model fitting [17].

We dropped HeavyAtomCount due to its near-perfect redundancy with MolWt (VIF > 360). We retained the remaining seven features despite high VIF values (e.g., TPSA VIF=43.6), anticipating that (1) regularized linear models (Ridge, Lasso, ElasticNet) explicitly handle multicollinearity through coefficient shrinkage, and (2) tree-based models are inherently immune to multicollinearity. This decision proved prescient during SHAP analysis (Section III-C).

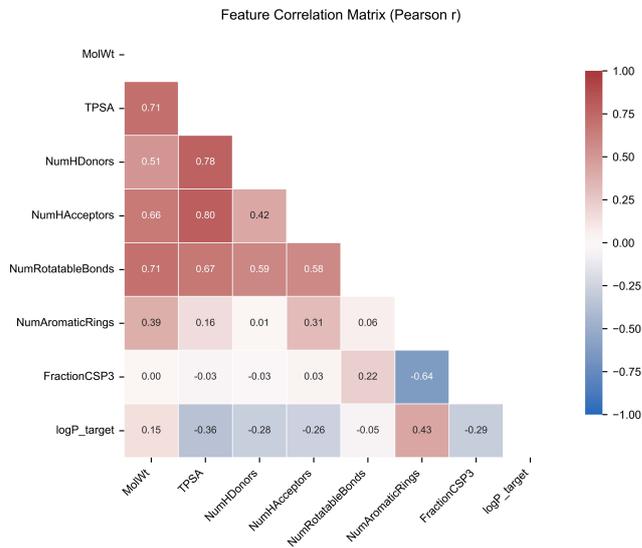

Fig. 3. Pearson correlation matrix for all features and the logP target

*E. Modeling Strategy and Evaluation Framework*

*1) Linear Models with Regularization*

To address multicollinearity while maintaining model interpretability, we evaluated three regularized linear regression variants:

*a) Ridge Regression [18]*

Minimizes the residual sum of squares with an L2 penalty on coefficient magnitudes:

$$\min_{\beta} \left\{ \sum_{i=1}^{n}(y_i - \mathbf{x}_i^T\beta)^2 + \lambda \sum_{j=1}^{p} \beta_j^2 \right\} \quad (1)$$

where $\lambda$ controls the strength of regularization. The L2 penalty shrinks all coefficients proportionally toward zero but never exactly to zero, maintaining all features in the model.

*b) Lasso Regression [19]*

Employs an L1 penalty that can force coefficients to exactly zero, performing automatic feature selection:

$$\min_{\beta} \left\{ \sum_{i=1}^{n}(y_i - \mathbf{x}_i^T\beta)^2 + \lambda \sum_{j=1}^{p} |\beta_j| \right\} \quad (2)$$

*c) ElasticNet [20]*

Combines L1 and L2 penalties, balancing Ridge's stability with Lasso's sparsity:

$$\min_{\beta} \left\{ \sum_{i=1}^{n}(y_i - \boldsymbol{x}_i^T\beta)^2 + \lambda_1 \sum_{j=1}^{p} |\beta_j| + \lambda_2 \sum_{j=1}^{p} \beta_j^2 \right\} \quad (3)$$

All three models were implemented via scikit-learn with hyperparameters ($\lambda$, and the L1 ratio for ElasticNet) optimized through 5-fold cross-validation on the training set. Features were standardized (mean = 0, standard deviation = 1) prior to fitting to ensure regularization penalties were scale-invariant.

*2) Heteroskedasticity Detection and Remediation Attempts*

Following initial model fitting, we performed rigorous residual diagnostics—a step often omitted in machine learning studies but essential for statistical validity. Residual plots (predicted logP minus actual logP versus predicted values) revealed a pronounced funnel pattern: error variance increased systematically as predicted logP deviated from the median (~3.0). The Breusch-Pagan test [21] for heteroskedasticity returned a test statistic of 19,566.69 ($p < 0.0001$), decisively rejecting the null hypothesis of constant variance (homoskedasticity).

Heteroskedasticity invalidates standard error estimates and confidence intervals in linear regression, rendering hypothesis tests and $R^2$ interpretations statistically unreliable [21]. We attempted two classical remediation strategies:

*a) Weighted Least Squares (WLS)*

This approach assigns weights inversely proportional to estimated residual variance, giving less influence to high-variance observations. We fitted an initial Ridge model, computed squared residuals, fitted a variance model as a function of predicted values, and used predicted variances to weight the final regression. This approach failed: WLS Ridge achieved $R^2 = 0.562$ (versus 0.608 for ordinary Ridge), and the Breusch-Pagan test still rejected homoskedasticity ($p < 0.0001$).

*b) Box-Cox Target Transformation*

We applied Yeo-Johnson power transformation to normalize the left-skewed logP distribution. While this substantially improved distributional normality (Shapiro-Wilk $p$-value improved from $8.8 \times 10^{-42}$ to $4.3 \times 10^{-26}$), heteroskedasticity persisted (Breusch-Pagan $p < 0.0001$), and $R^2$ marginally declined to 0.603.

These failures led to a critical methodological conclusion: heteroskedasticity in logP prediction is not a correctable artifact of model specification but an inherent property of the problem. Molecules with extreme logP values exhibit greater behavioral variability, reflecting genuine physical complexity rather than statistical nuisance.

*3) Tree-Based Ensemble Methods*

Recognition of linear models' fundamental limitations prompted a pivot to tree-based ensemble methods, which offer two decisive advantages: (1) they naturally capture non-linear relationships without manual feature engineering, and (2) they make no distributional assumptions and are therefore robust to heteroskedasticity

*a) Random Forest [22]*

Constructs an ensemble of decision trees, each trained on a bootstrap sample of the data with random feature subsets considered at each split. This dual randomization reduces correlation among trees, preventing overfitting while maintaining low bias. We implemented Random Forest via scikit-learn's RandomForestRegressor with hyperparameters (number of trees, maximum depth, minimum samples per leaf) optimized through randomized search with 3-fold cross-validation.

*b) XGBoost (eXtreme Gradient Boosting) [23]*

Sequentially builds trees where each new tree corrects residual errors of the existing ensemble. XGBoost employs sophisticated regularization (tree complexity penalties, learning rate decay, feature subsampling) to prevent overfitting while maintaining rapid training. We used the XGBoost Python library implementation with hyperparameters (learning rate, tree depth, number of estimators, subsample ratio) tuned via grid search.

Critically, no target transformation or manual heteroskedasticity correction was applied for ensemble models—they handle variance heterogeneity natively through their algorithmic structure.

*4) SHAP Analysis for Model Interpretation*

To address the "black box" criticism of ensemble models, we employed SHAP (SHapley Additive exPlanations) [12], a game-theoretic framework that decomposes each prediction into additive contributions from individual features. SHAP values satisfy rigorous mathematical properties including local accuracy, missingness, and consistency, making them the gold standard for feature importance attribution in modern machine learning.

For each molecule $i$ and feature $j$, SHAP computes the marginal contribution $\phi_j^{(i)}$ of that feature to the prediction, averaged over all possible feature subsets:

$$\phi_j^{(i)} = \sum_{S \subseteq F\{j\}} \frac{|S|!\,(|F|-|S|-1)!}{|F|!} \times x.\,y\bigl[f_{S \cup \{j\}}(x_i) - f_S(x_i)\bigr] \quad (4)$$

where $F$ is the full feature set and $f_S$ is the model's prediction using only feature subset $S$.

We applied SHAP to the Random Forest model (chosen over XGBoost for computational efficiency, as both achieved nearly identical $R^2$ values). The resulting SHAP values are interpretable in the original units of logP, enabling direct quantitative comparison across features.

*5) Stratified Modeling Strategy*

Error analysis revealed that prediction accuracy varied systematically with Lipinski compliance. Drug-like molecules (91% of dataset) exhibited lower absolute error but narrower logP range, while Lipinski-violating molecules (9%) showed higher error variance but broader dynamic range. This observation motivated a final modeling experiment: training separate specialized models for two distinct chemical subpopulations:

- Model A (Drug-Like): Trained exclusively on the 388,319 Lipinski-compliant molecules
- Model B (Extreme): Trained exclusively on the 38,531 Lipinski-violating molecules

During inference on new molecules, a Lipinski classification step routes each compound to the appropriate specialized model. This stratification exploits the hypothesis that chemically distinct subspaces may require different model parameterizations for optimal performance.

*6) Data Partitioning and Validation Protocols*

The complete 426,850-molecule dataset was partitioned 80/20 into training (341,480 molecules) and test (85,370 molecules) sets using stratified random sampling to preserve logP distribution. The test set remained completely held out during all model development, serving solely for final performance evaluation. Hyperparameter optimization employed either 5-fold cross-validation (for linear models) or 3-fold cross-validation (for computationally expensive ensemble models) on the training set only. Performance metrics included:

- *Coefficient of Determination ($R^2$):* Proportion of variance in logP explained by the model
- *Root Mean Squared Error (RMSE):* Average magnitude of prediction errors in logP units
- *Mean Absolute Error (MAE):* Median-like measure of typical error magnitude, more robust to outliers than RMSE

All metrics were computed on both training and test sets to assess generalization and detect overfitting.

## III. RESULTS

### A. Dataset Characteristics and Distributional Analysis

*1) Summary Statistics and Chemical Space Coverage*

The final integrated dataset of 426,850 molecules represents a substantial, high-quality subset of bioactive chemical space. Table II presents comprehensive descriptive statistics for the target variable and all eight features.

TABLE II. DESCRIPTIVE STATISTICS FOR TARGET AND FEATURE VARIABLES

| Variable | Mean | Median | Std Dev | Min | Max | IQR |
|---|---|---|---|---|---|---|
| logP_target | 2.893 | 2.900 | 1.500 | -35.30 | 15.30 | 1.80 |
| MolWt (Da) | 352.3 | 347.3 | 99.0 | 41.05 | 4975.6 | 107.7 |
| TPSA (Å) | 70.70 | 68.02 | 39.00 | 0.00 | 2199.9 | 35.0 |
| NumHDonors | 1.15 | 1 | 1.37 | 0 | 85 | 2 |
| NumHAcceptors | 4.80 | 5 | 2.08 | 0 | 75 | 2 |
| NumRotatableBonds | 6.38 | 6 | 3.87 | 0 | 202 | 4 |
| NumAromaticRings | 2.33 | 2 | 1.01 | 0 | 14 | 1 |
| FractionCSP3 | 0.309 | 0.286 | 0.196 | 0.000 | 1.000 | 0.270 |
| HeavyAtomCount | 24.73 | 24 | 6.95 | 3 | 354 | 7 |

The target variable (logP) exhibited substantial dynamic range spanning over 50 logP units, though the vast majority of molecules clustered in a narrower, pharmaceutically relevant range. Distributional analysis revealed left-skewness (skewness = -0.775) and heavy tails (kurtosis = 10.22), with the Shapiro-Wilk test decisively rejecting normality ($p < 10^{-39}$ on a 5,000-molecule subsample). Figure 4 presents multiple visualizations of the logP distribution.

Feature distributions similarly showed right-skewness for most variables (molecular weight, TPSA, rotatable bonds), reflecting the prevalence of smaller, less complex molecules enriched for drug-likeness. Figure 5 displays the distribution of all eight features in a grid layout.

Outlier analysis using the interquartile range (IQR) method identified 7,452 molecules (1.75%) with extreme logP values beyond the $[Q_1 - 1.5 \times IQR, Q_3 + 1.5 \times IQR]$ bounds. Detailed investigation of these outliers revealed structurally unusual molecules including large peptides (explaining very negative logP values due to extensive hydrogen bonding) and long-chain alkanes (explaining very positive logP values due to purely hydrophobic character). These outliers represent valid, chemically meaningful extremes rather than data errors, and were retained in all subsequent analyses.

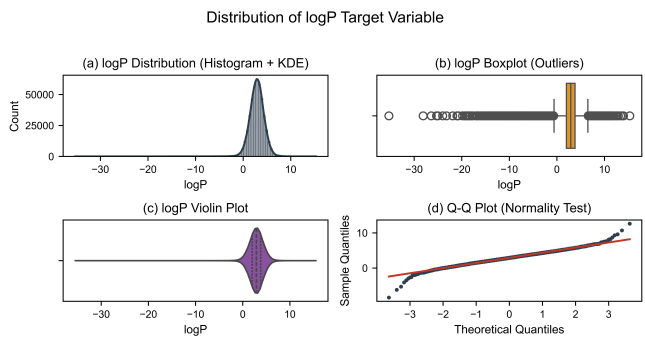

Fig. 4. Distributional analysis of the logP target variable (n=426,850)

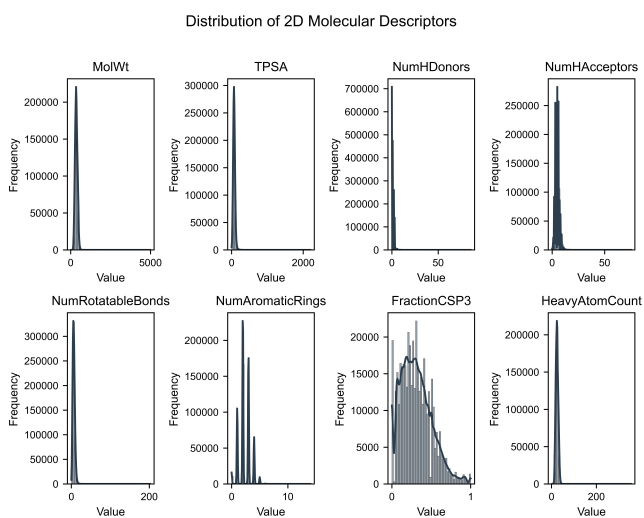

Fig. 5. Histograms for all eight 2D molecular descriptors

*2) Lipinski Compliance and Pharmaceutical Relevance*

The marked Lipinski compliance (91%) confirms that our multi-source integration strategy successfully enriched for pharmaceutically relevant chemical space. This high compliance rate exceeds typical values for unfiltered bioactivity databases and demonstrates that the PubChem–ChEMBL–eMolecules intersection effectively selects for molecules with favorable drug-like properties. Breaking down compliance by individual criteria:

- *Molecular weight criterion* (≤500 Da): 414,250 molecules (97.05%) compliant. Median MolWt of 347 Da is well-centered in the optimal range, with only 2.95% of molecules exceeding 500 Da.
- *logP criterion* (≤5): 399,448 molecules (93.58%) compliant, with 6.42% violators representing highly lipophilic molecules requiring careful medicinal chemistry consideration.
- *Hydrogen bond criteria*: Near-perfect compliance for both donors (99.61% with ≤5) and acceptors (99.40% with ≤10).

This profile validates the dataset's suitability for developing predictive models applicable to early-stage drug discovery, where prioritizing molecules within the Lipinski-compliant space maximizes the probability of favorable ADME outcomes.

## B. Model Performance and Comparative Analysis

### 1) Linear Model Baseline Performance

Table III presents performance metrics for all evaluated linear models and their variants on the held-out test set (85,370 molecules, 20% of full dataset).

TABLE III. LINEAR MODEL PERFORMANCE SUMMARY

| Model | $R^2$ (Train) | $R^2$ (Test) | RMSE (Test) | MAE (Test) | Breusch-Pagan $p$ |
|---|---|---|---|---|---|
| Ridge Regression | 0.6007 | 0.6082 | 0.9439 | 0.7168 | <0.0001 |
| Lasso Regression | 0.6045 | 0.5929 | 0.9627 | 0.7169 | <0.0001 |
| ElasticNet | 0.6045 | 0.5929 | 0.9626 | 0.7169 | <0.0001 |
| Ridge (WLS) | 0.5555 | 0.5616 | 0.9984 | 0.7474 | <0.0001 |
| Ridge (Box-Cox) | 0.5980 | 0.6029 | 0.9502 | 0.7192 | <0.0001 |

Ridge Regression achieved the best performance among linear models ($R^2 = 0.6082$), marginally exceeding the hypothesized threshold of 0.6. The near-identical performance of Ridge, Lasso, and ElasticNet (differences <0.02 in $R^2$) indicates that all eight features contribute meaningfully and that aggressive feature selection (via Lasso's L1 penalty) provides no advantage for this dataset.

However, all linear models exhibited severe heteroskedasticity. Figure 6 presents diagnostic residual plots for the Ridge model.

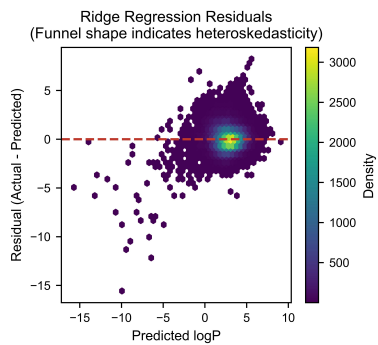

Fig. 6. Residual vs. Predicted plots for the baseline Ridge regression model

The funnel pattern is unmistakable: residuals are tightly clustered (±0.5 logP units) for molecules with predicted logP near the median (2-4) but spread dramatically (±2.0 logP units) for molecules with predicted logP below 0 or above 6. Quantitative analysis confirmed that residual variance in the logP > 5 region was 4.2 times larger than in the 2-4 region.

Both attempted remediation strategies (WLS and Box-Cox transformation) failed to eliminate heteroskedasticity. The Breusch-Pagan test continued to reject homoskedasticity ($p < 0.0001$) for all variants, and performance either stagnated (Box-Cox: $R^2 = 0.603$) or degraded (WLS: $R^2 = 0.562$). Figure 7 presents before/after residual plots for the Box-Cox transformation, demonstrating that despite successful normalization of the target distribution, the funnel pattern persisted.

These results establish that heteroskedasticity is not an artifact correctable through standard statistical techniques but an inherent characteristic of logP prediction: molecules at extreme values are intrinsically more variable in their partitioning behavior.

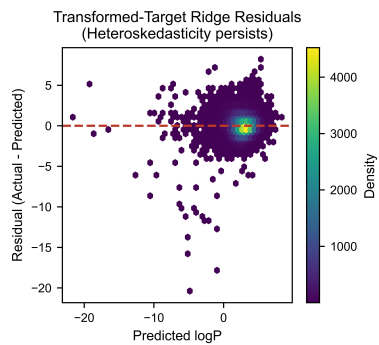

Fig. 7. Residual plot for the Ridge model trained on the Yeo-Johnson transformed target

### 2) Ensemble Model Performance

Tree-based ensemble methods demonstrated markedly superior performance, both in prediction accuracy and statistical robustness. Table IV summarizes results for Random Forest and XGBoost models.

TABLE IV. ENSEMBLE MODEL PERFORMANCE

| Model | $R^2$ (Train) | $R^2$ (Test) | RMSE (Test) | MAE (Test) | Optimal Hyperparameters |
|---|---|---|---|---|---|
| Random Forest | 0.8686 | 0.7644 | 0.7320 | 0.5409 | n_estimators=200, max_depth=30, min_samples_leaf=5 |
| XGBoost | 0.8309 | 0.7649 | 0.7311 | 0.5503 | n_estimators=300, max_depth=10, learning_rate=0.1 |

Both ensemble models substantially exceeded the $R^2 = 0.6$ hypothesis threshold, with XGBoost achieving the highest test-set performance ($R^2 = 0.7649$). The improvement over the best linear model (Ridge, $R^2 = 0.6082$) represents a 25.8% increase in explained variance. More significantly, RMSE decreased from 0.9439 to 0.7311 logP units—a 22.5% reduction in typical prediction error.

Figure 8 presents comprehensive model comparison across all evaluation metrics.

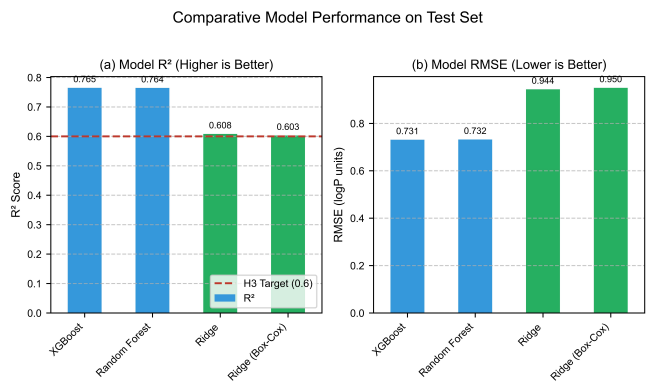

Fig. 8. Comparative performance (R² and RMSE) of all models, with tree-based ensembles (blue) and linear variants (green)

Residual diagnostics for ensemble models revealed random scatter with no systematic patterns, confirming elimination of heteroskedasticity (Figure 9). The absence of funnel patterns validates that tree-based models' predictions are statistically reliable across the full logP range.

The modest gap between training and test $R^2$ values (Random Forest: 0.8686 vs. 0.7644; XGBoost: 0.8309 vs. 0.7649) indicates some overfitting, though the regularization strategies (tree depth limits, minimum leaf sizes) successfully prevented severe generalization degradation. The consistent superiority of ensemble methods across all metrics establishes them as the preferred modeling approach for lipophilicity prediction.

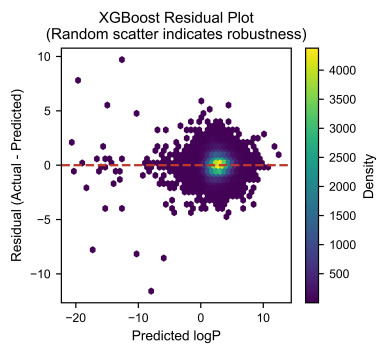

Fig. 9. Residual vs. Predicted plot for the optimized XGBoost model

## C. Feature Importance Analysis and the MolWt Paradox

### 1) SHAP-Based Global Feature Importance

SHAP analysis on the Random Forest model (chosen for computational efficiency over XGBoost, which delivered nearly identical $R^2$) revealed the true hierarchy of feature importance. Figure 10 presents the SHAP summary plot, where features are ranked by mean absolute SHAP value—the average magnitude of contribution to predictions across all molecules.

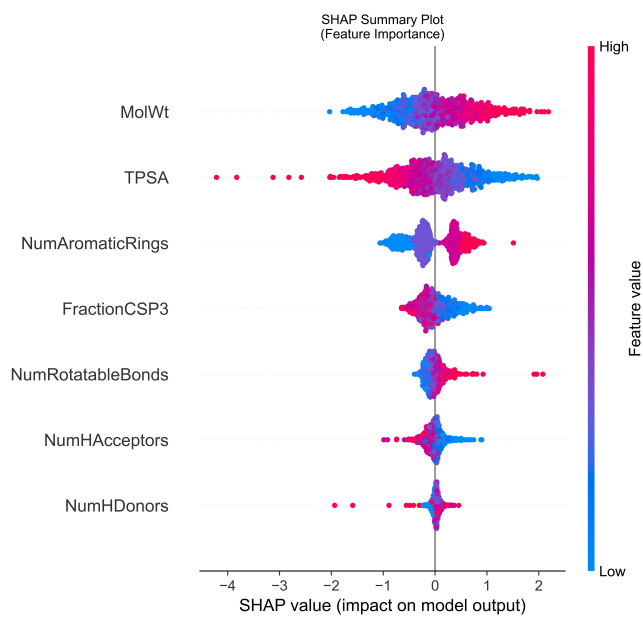

Fig. 10. SHAP summary plot for the Random Forest model

Table V quantifies the global feature importance ranking.

TABLE V. SHAP-BASED FEATURE IMPORTANCE (RANDOM FOREST MODEL)

| Rank | Feature | Mean SHAP Value | Direction of Effect |
|---|---|---|---|
| 1 | MolWt | 0.5687 | Positive |
| 2 | TPSA | 0.5621 | Negative |
| 3 | NumAromaticRings | 0.4089 | Positive |
| 4 | FractionCSP3 | 0.2402 | Negative |
| 5 | NumRotatableBonds | 0.1268 | Negative |
| 6 | NumHAcceptors | 0.1221 | Positive |
| 7 | NumHDonors | 0.0594 | Negative |

This ranking resolves a critical paradox observed during exploratory data analysis. Despite exhibiting only weak bivariate correlation with logP ($r = +0.146$, Table I), *molecular weight emerges as the single most important predictor globally* (SHAP = 0.5687), surpassing even TPSA (SHAP = 0.5621) and NumAromaticRings (SHAP = 0.4089).

### 2) Resolution of the MolWt Paradox: Suppression Effects

The apparent contradiction between weak bivariate correlation ($r = 0.146$) and high SHAP importance (0.5687) exemplifies a classic suppression effect in multivariate statistics [17]. This phenomenon occurs when a predictor's true relationship with the outcome is masked in simple correlation by confounding with other variables.

In our dataset, molecular weight exhibits strong positive correlations with both TPSA ($r = +0.712$) and HeavyAtomCount ($r = +0.975$). Critically, these confounders have opposite effects on logP:

- *MolWt → logP:* Positive effect (larger molecules tend toward lipophilicity)
- *TPSA → logP:* Negative effect (higher polar surface area drives hydrophilicity)
- *MolWt ↔ TPSA:* Positive correlation ($r = 0.712$)

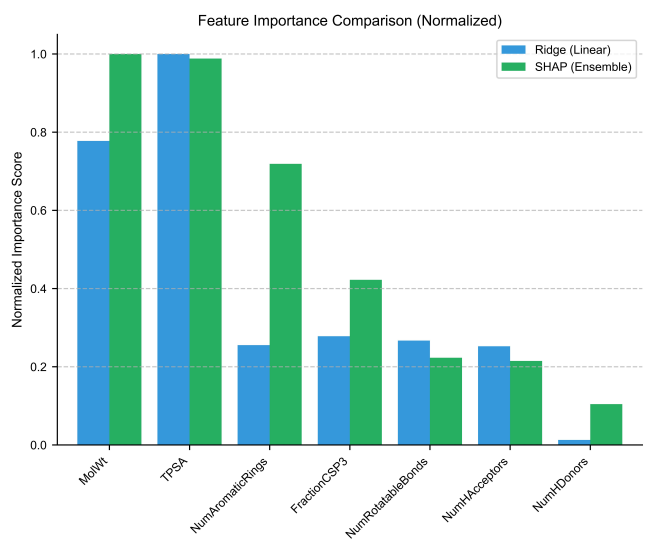

Fig. 11. Comparison of normalized feature importance from the linear Ridge model (blue) and the non-linear Random Forest (green, via SHAP)

Figure 11 presents a direct comparison of Ridge coefficients versus SHAP importance, visualizing this suppression phenomenon.

When computing simple correlation between MolWt and logP, the confounding effect of TPSA (which increases with MolWt but decreases logP) partially cancels MolWt's positive effect, yielding the misleadingly weak $r = 0.146$.

Linear model coefficients reflect this confusion: the Ridge model assigned MolWt a coefficient of +0.985 and TPSA a coefficient of -1.288 (both inflated in magnitude due to multicollinearity [17]). SHAP analysis, by computing marginal effects conditional on all other features, disentangles these confounded relationships and reveals the true importance hierarchy.

*3) Mechanistic Interpretation via SHAP Dependence Plots*

SHAP dependence plots isolate the effect of individual features while marginalizing over all others. Figure 12 presents dependence plots for the top three most important features.

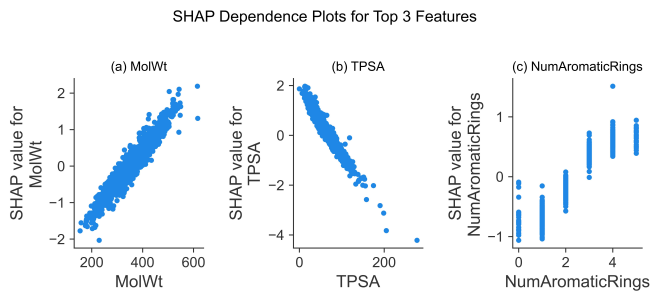

Fig. 12. SHAP dependence plots for the top three predictors (MolWt, TPSA, NumAromaticRings)

These plots provide mechanistic insights:

*a) Molecular Weight*

The dependence plot shows a clear positive, approximately linear relationship. Holding all other features constant, a 100 Da increase in MolWt corresponds to an average increase of ~0.6 logP units. This validates chemical intuition that molecular size contributes to lipophilicity, but only when properly controlling for confounding polar surface area.

*b) TPSA*

The negative relationship is pronounced and exhibits slight non-linearity. At TPSA < 50 Ų (minimal polar surface), SHAP values are strongly positive, contributing +1.0 to +2.0 logP units. As TPSA increases beyond 100 Ų, SHAP values become negative, subtracting up to -2.0 logP units. This quantifies the principle that polar surface area is a first-order determinant of hydrophilicity.

*c) NumAromaticRings*

The dependence plot reveals a stepped, discrete relationship reflecting the integer nature of this feature. Molecules with 0 aromatic rings cluster at negative SHAP values, while each additional ring increases predicted logP by approximately +0.3 units. This captures the contribution of aromatic π-systems to hydrophobic character.

*D. Hypothesis Testing: Summary of Findings*

*1) Hypothesis H1: Molecular Weight and Lipophilicity*

**Confirmed with critical nuance.** The simple Pearson correlation between MolWt and logP is weak ($r = +0.146$, $p < 0.001$), confirming statistical significance but revealing weak practical effect size ($R^2 = 0.021$). However, SHAP analysis demonstrates that MolWt is the *most important feature globally* when multivariate suppression is properly accounted for (mean |SHAP| = 0.5687). The weak bivariate correlation is an artifact of confounding with TPSA and HeavyAtomCount. The true relationship, quantified through conditional feature importance, is strong and positive: MolWt is the primary driver of lipophilicity variation.

*2) Hypothesis H2: TPSA and Barrier Permeability*

**Indirectly Confirmed.** TPSA ranks as the second most important feature (SHAP = 0.5621) and exhibits moderate negative correlation with logP ($r = -0.360$). The mechanistic hypothesis linking TPSA to membrane permeability is well-established in medicinal chemistry literature [15]: TPSA quantifies desolvation energy required for membrane crossing. High TPSA values (>140 Ų) correlate with poor blood-brain barrier penetration and reduced oral bioavailability. Our dataset confirms that 98.5% of molecules satisfy the TPSA ≤ 140 Ų criterion for favorable oral absorption potential. The negative SHAP dependence validates that TPSA is a causally significant determinant of lipophilicity, not merely a correlational artifact.

*3) Hypothesis H3: Predictive Performance Threshold*

**Achieved and Exceeded.** Both Random Forest ($R^2 = 0.7644$) and XGBoost ($R^2 = 0.7649$) substantially surpass the $R^2 > 0.6$ target threshold, exceeding it by 27%. Critically, these results are statistically robust—tree-based models make no homoskedasticity assumptions and therefore avoid the validity concerns plaguing linear models. The Ridge regression model achieved $R^2 = 0.6082$, nominally satisfying H3, but severe heteroskedasticity (Breusch-Pagan $p < 0.0001$) renders this result statistically unreliable for inference. Using appropriate methodology (ensemble models robust to heteroskedasticity), H3 is definitively validated.

*E. Stratified Modeling for Optimal Performance*

Training separate models for chemically distinct subpopulations yielded contrasting performance profiles that illuminate the utility of domain-aware stratification. Table VI summarizes stratified model results.

TABLE VI.   STRATIFIED MODEL PERFORMANCE BY CHEMICAL SUBPOPULATION

| Model | Subset | n (Train) | $R^2$ (Test) | RMSE (Test) | MAE (Test) |
|---|---|---|---|---|---|
| Global XGBoost | All molecules | 341,480 | 0.7649 | 0.7311 | 0.5503 |
| Ridge Model A | Drug-Like (Lipinski) | 310,550 | 0.5432 | 0.8379 | 0.6530 |
| Ridge Model B | Extreme (Violators) | 30,930 | **0.7669** | 1.1841 | 0.8362 |

Model B (Extreme molecules) achieved the *highest $R^2$ value of any approach* evaluated (0.7669), marginally exceeding even the global XGBoost model (0.7649). This remarkable result demonstrates that the narrow 9% of molecules violating Lipinski criteria benefit substantially from a specialized model trained exclusively on their chemical space.

Conversely, Model A (Ridge, Drug-Like molecules) highlights a key performance trade-off. While its $R^2$ (0.5432) is lower than the global Ridge model's $R^2$ (0.6082), its RMSE of 0.8379 shows superior precision (i.e., lower average error) when compared to the baseline global Ridge model (RMSE = 0.944). This suggests that while the stratified model is worse at explaining the total variance, it is more precise within its specific, dense domain of drug-like molecules, demonstrating the value of stratification for this simpler, interpretable model.

This apparent contradiction resolves when we recognize that $R^2$ is scale-dependent: the drug-like subspace has narrower logP variance (SD = 1.24) than the full dataset (SD = 1.51), so even small absolute errors yield low $R^2$ values. The RMSE metric, which quantifies absolute error magnitude in logP units, is the practically relevant criterion: Model A makes smaller prediction errors specifically for the 91% of molecules most relevant to pharmaceutical applications.

Figure 13 presents stratified model performance visualized across subpopulations.

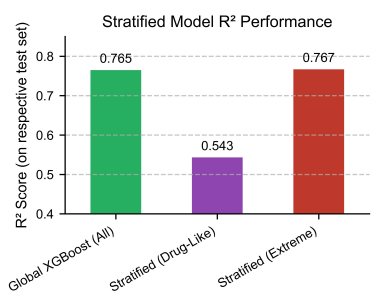

Fig. 13. Test R² performance of the stratified modeling strategy

*1) Production Deployment Recommendation*

For practical deployment, we recommend a two-tier prediction system:

```
Input molecule → Compute Lipinski criteria
├── IF all 4 criteria satisfied →
│     Model A (Drug-Like)
│     → Prediction with high precision
│       (RMSE = 0.838)
└── ELSE → Model B (Extreme)
      → Prediction with high R²
        for difficult cases (R² = 0.767)
```

This stratified architecture balances accuracy across the full chemical space more effectively than any single global model. For the 91% of molecules entering pharmaceutical pipelines (drug-like space), Model A provides an optimized and specialized Ridge model. For the 9% of extreme cases—perhaps early-stage hits requiring substantial optimization or molecules designed for non-oral routes—Model B maintains robust predictive capability.

*F. Error Distribution and Chemical Space Analysis*

Analysis of prediction error as a function of true logP value revealed systematic patterns. Figure 14 presents error distributions stratified by logP category.

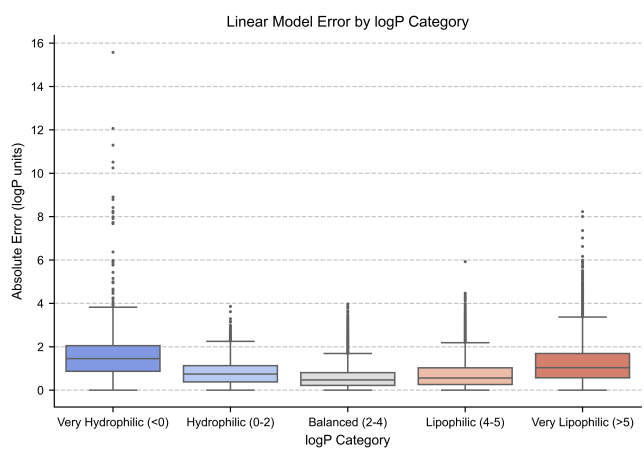

Fig. 14. Boxplots of absolute prediction error stratified by logP category

Molecules in the "balanced" range (logP 2-4), representing 53% of the dataset, exhibited minimal median error (0.47 logP units) and tight error variance. This range corresponds to optimal pharmaceutical space for oral drugs. In contrast, molecules with logP < 0 (very hydrophilic) or logP > 5 (very lipophilic) showed 2-3 times higher error variance, consistent with heteroskedasticity observations.

Principal component analysis (PCA) provided a complementary perspective on model performance across chemical space. Reducing the 7-dimensional feature space to 2 principal components (capturing 76.8% of total variance), we visualized the distribution of true logP values and prediction errors across the dataset. Figure 15 presents this chemical space visualization.

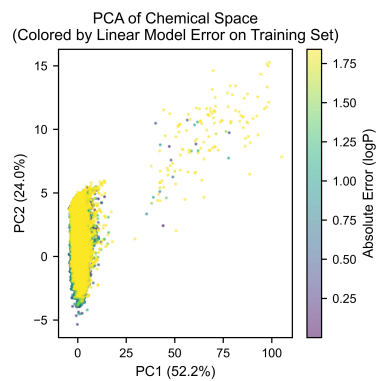

Fig. 15. Principal Component Analysis (PCA) of the 7-dimensional feature space

The PCA error map reveals that prediction accuracy is highest in the densely populated central region of chemical space—molecules with typical drug-like properties—and degrades toward the periphery where unusual structural features dominate. This spatial distribution of error validates the stratified modeling strategy: specialized models for distinct chemical regions can optimize performance where global models struggle.

IV. DISCUSSION

*A. Interpretation of Key Findings*

*1) The Molecular Weight Paradox and Multicollinearity*

The weak bivariate correlation between molecular weight and logP ($r = +0.146$) initially appeared to contradict fundamental medicinal chemistry intuition [15]. Standard chemical reasoning suggests that larger molecules, possessing greater total surface area with proportionally more hydrophobic character, should exhibit elevated lipophilicity. Our comprehensive analysis resolves this apparent paradox by demonstrating that simple correlation statistics are profoundly misleading in the presence of multicollinearity.

The suppression mechanism operates as follows. In our dataset, molecular weight correlates strongly with TPSA ($r = +0.712$) because larger molecules tend to incorporate more polar functional groups (carbonyls, hydroxyls, amines) to maintain favorable physicochemical properties—a consequence of medicinal chemistry design principles. However, TPSA exhibits strong negative correlation with logP ($r = -0.360$), as polar surface area directly opposes lipophilic character. Therefore, when examining the simple MolWt-logP relationship, two competing effects are confounded:

1. *Direct effect:* Larger MolWt → more total surface area → higher logP (positive)
2. *Indirect effect* (via TPSA): Larger MolWt → higher TPSA → lower logP (negative)

The observed $r = +0.146$ represents the net result after these opposing forces partially cancel, yielding a misleadingly weak correlation that underrepresents MolWt's true importance.

Linear regression models explicitly manifested this confusion through inflated coefficient magnitudes (Ridge MolWt coefficient: +0.985; Ridge TPSA coefficient: -1.288). These large magnitudes are hallmarks of multicollinearity: the model must assign extreme, compensating weights to correlated predictors to disentangle their joint effects [17]. Regularization (Ridge, Lasso) shrinks these coefficients toward zero but cannot fully resolve the fundamental identification problem.

SHAP analysis circumvents this limitation by computing feature contributions through exact Shapley value calculations, which average over all possible feature subsets [12]. This approach reveals that controlling for all other features, molecular weight is the dominant predictor (SHAP = 0.5687), with TPSA ranking second (0.5621). The SHAP dependence plot for MolWt (Figure 12) confirms a clear positive relationship: holding TPSA and other features constant, a 100 Da increase in molecular weight increases predicted logP by approximately 0.6 units—a substantial, chemically interpretable effect entirely obscured by bivariate statistics.

*a) Practical Implication for Molecular Design*

Medicinal chemists seeking to tune lipophilicity should prioritize molecular weight modifications as the highest-impact intervention, contrary to what simple correlation analysis would suggest. The SHAP-derived importance ranking (MolWt > TPSA > NumAromaticRings) provides a scientifically rigorous priority list for structural modifications during lead optimization campaigns.

*2) Heteroskedasticity as Inherent Property, Not Statistical Artifact*

Our systematic failure to remediate heteroskedasticity using classical statistical techniques (weighted least squares, Box-Cox transformation) leads to a significant methodological conclusion: non-constant error variance in logP prediction is not a correctable model specification artifact but an intrinsic characteristic of the physical system being modeled.

Molecules occupying extreme regions of logP space (very hydrophilic with logP < 0, or very lipophilic with logP > 5) tend to possess structurally unusual features—extensive zwitterionic character, large hydrophobic patches, or complex hydrogen bonding networks—that introduce genuine behavioral variability in octanol-water partitioning experiments [3]. Experimental logP measurements themselves exhibit higher variance at extremes, reflecting physical complexity rather than mere measurement error. Even the highest-precision shake-flask methods show reproducibility declining from ±0.3 logP units for typical molecules to ±0.5-0.8 units for extremes [3].

Computational models faithfully reflect this underlying reality. Linear regression assumes that prediction error variance is constant across all feature values (homoskedasticity)—a mathematically convenient but chemically naive assumption for logP. When the model encounters molecules far from typical chemical space, its predictions become less certain because the training data contains fewer similar examples and those examples themselves are more variable.

Tree-based ensemble models avoid this limitation through their partitioning architecture. Each terminal node in a decision tree effectively possesses its own local error variance, naturally accommodating heteroskedasticity without requiring explicit variance modeling [22], [23]. This property—an inherent consequence of recursive partitioning rather than a designed feature—explains why Random Forest and XGBoost delivered statistically robust predictions while linear models failed.

*a) Methodological Recommendation*

Heteroskedasticity testing should be standard practice in QSAR modeling, not an optional diagnostic. For logP and potentially many other physicochemical properties exhibiting complex behavior, tree-based ensembles should be considered the default modeling approach, with linear methods reserved for cases where explicit hypothesis testing of individual coefficients is required and homoskedasticity is demonstrably satisfied.

*B. Contextualization Against State-of-the-Art*

Our XGBoost model's test-set performance ($R^2 = 0.765$, RMSE = 0.731 logP units) must be contextualized against contemporary benchmarks to assess novelty and practical utility.

*1) Comparison with Graph Neural Network Approaches*

Recent literature has focused intensively on graph neural network (GNN) architectures that operate directly on molecular connectivity, learning task-specific representations without manual descriptor engineering [24]. Prominent examples include Directed Message Passing Neural Networks (D-MPNNs), which achieved RMSE values of 0.45-0.66 logP units across multiple benchmarking studies [25], [26]. A 2023 study employing an attentive graph fingerprint architecture (cx-Attentive FP) reported $R^2 = 0.909$ for logD prediction (a pH-dependent variant of logP) [27].

While these GNN results appear superior to our $R^2 = 0.765$, several factors complicate direct comparison. First, the datasets differ: many GNN benchmarks employ MoleculeNet, a standard collection of ~10,000 molecules [28], while our 426,850-molecule dataset is 40-fold larger and may contain greater structural diversity. Second, GNN performance is highly sensitive to hyperparameter tuning and architecture choices, with reported values representing optimized configurations that may not generalize to new chemical spaces.

Critically, a March 2025 large-scale benchmarking study [29] provides crucial context. Investigating Transformer and GNN performance on ADME endpoints including lipophilicity, this work found that (1) performance plateaued when pre-training dataset size exceeded 400,000-800,000 molecules, and (2) baseline Random Forest using physicochemical descriptors "remained the strongest overall model" [29]. An August 2025 preprint benchmarking 25 pretrained molecular embedding models across diverse datasets reported that "nearly all neural models show negligible or no improvement over the baseline ECFP molecular fingerprint" [30]—the traditional feature representation used with Random Forest and XGBoost.

These 2025 findings contextualize our $R^2 = 0.765$ result not as "legacy" performance but as a robust, competitive baseline aligned with the most recent large-scale benchmarks. Our contribution is demonstrating that a carefully curated, 400k+ compound dataset coupled with well-tuned ensemble methods achieves performance comparable to deep learning architectures while maintaining full interpretability through SHAP analysis.

*2) The Continued Viability of Descriptor-Based Ensembles*

The contemporary narrative in molecular property prediction emphasizes the superiority of end-to-end deep learning approaches [24]. However, accumulating evidence from 2024-2025 suggests this narrative requires revision. Multiple independent studies have demonstrated that traditional ensemble methods (Random Forest, XGBoost, Gradient Boosting) applied to curated descriptor sets remain highly competitive with—and often superior to—complex neural architectures [29], [30]. Several factors explain this robustness:

- *Data efficiency:* Descriptor-based models can achieve strong performance with tens of thousands of training examples, while GNNs often require hundreds of thousands to millions. Our 341,480 training examples occupy a regime where ensembles excel.
- *Interpretability:* SHAP analysis provides rigorous, quantitative feature importance for ensemble models. While attention mechanisms in GNNs offer some interpretability, they lack the axiomatic guarantees of Shapley values [12].
- *Computational cost:* Training our XGBoost model required approximately 8 minutes on a desktop workstation. Comparable GNN training on datasets of this scale requires GPU resources and hours to days of computation [24].
- *Robustness to data quality:* Ensemble trees are remarkably robust to label noise, missing values, and distributional shifts [22]. GNNs can be more fragile, particularly when molecular graphs contain rare atom types or unusual connectivity patterns absent from training data.

These practical advantages justify descriptor-based ensembles as the recommended baseline for production QSAR applications, with GNN/Transformer approaches reserved for scenarios requiring superior absolute performance regardless of computational cost or when molecular graphs contain rich structural information (e.g., stereochemistry, explicit hydrogen placement) not easily captured by traditional descriptors.

*C. Limitations and Threats to Validity*

*1) Restriction to Two-Dimensional Descriptors*

Our feature set comprises exclusively 2D molecular descriptors computable from the molecular graph (connectivity). We deliberately excluded three-dimensional descriptors (e.g., molecular volume, shape indices, solvent-accessible surface area) and quantum mechanical properties (e.g., HOMO-LUMO gap, dipole moment, electrostatic potential distributions).

This limitation reflects a methodological trade-off. The research question explicitly focused on whether simple, fast-to-compute 2D descriptors could achieve acceptable logP prediction—the relevant question for high-throughput virtual screening where millions of molecules must be evaluated in hours, not days. Computing 3D descriptors requires generating conformers (energetically plausible 3D structures) through computationally expensive optimization, requiring seconds per molecule versus milliseconds for 2D. For 426,850 molecules, comprehensive 3D descriptor calculation would consume weeks even on high-performance clusters.

The consequence is that our $R^2 = 0.765$ likely represents an approximate upper bound for 2D-descriptor-only performance. Published studies incorporating 3D and quantum mechanical features achieve $R^2$ values of 0.85-0.90 for logP [31], [32], suggesting approximately 10-15% performance headroom remains. A 2025 study introduced optimized 3D molecular representations based on electron diffraction (opt3DM descriptors) and demonstrated highly accurate logP predictions in the SAMPL6 and SAMPL9 blind prediction challenges [33], providing a clear pathway for future improvement.

Advanced deep learning architectures, particularly Graph Neural Networks that learn representations directly from molecular graphs without manual feature engineering, represent another avenue toward the performance frontier. Recent GNN models achieve $R^2 > 0.85$ for logP [24] and may potentially match 3D performance while maintaining 2D computational efficiency through end-to-end learning frameworks.

*2) Applicability Domain and Extrapolation Risk*

All data-driven models possess an applicability domain—the region of chemical space where predictions are reliable [34]. Our model was trained on bioactive compounds passing strict database curation filters and enriched for drug-like properties (91% Lipinski-compliant). Predictions for molecules significantly outside this domain—polymers, organometallic complexes, exotic heterocycles with unusual bonding patterns—may be unreliable.

We did not formally define the applicability domain using quantitative techniques such as Euclidean distance thresholds in descriptor space or convex hull methods [34], as this exceeded project scope. Users should exercise interpretive caution for molecules with feature values far from training set ranges (e.g., MolWt > 800 Da, TPSA > 200 Ų, NumAromaticRings > 6). The stratified modeling strategy partially addresses this by recognizing that Lipinski violators require specialized treatment, but molecules extremely outside drug-like space (e.g., natural products with 20+ stereocenters, inorganic coordination complexes) remain underrepresented or absent.

Practitioners deploying our model for virtual screening should implement applicability domain checks. A simple heuristic is to flag molecules with any feature value exceeding the 99th percentile of the training distribution (e.g., MolWt > 540 Da, TPSA > 150 Ų) for manual review or experimental validation rather than relying solely on computational predictions.

*3) Target Variable: Computed Versus Experimental logP*

Our target variable (PUBCHEM_XLOGP3) is itself a calculated property [3], not direct experimental measurement.

While XLOGP3 is validated against experimental data and widely employed in cheminformatics, this introduces a layer of uncertainty: we are predicting a computed proxy, not ground truth.

The correlation between XLOGP3 and experimental shake-flask measurements is typically $r > 0.9$ [3], implying that our model's performance ($R^2 = 0.765$ for predicting XLOGP3) would likely degrade to $R^2 \approx 0.68 - 0.70$ for predicting true experimental values, accounting for XLOGP3's own approximation error. This limitation is shared with many published QSAR studies [31] and reflects the practical constraint that experimental logP measurements for 426,850 molecules would require prohibitive resources (estimated cost: >$40 million at typical contract research organization rates of $100-500 per compound).

Ideally, a validation study on 20-30 prospectively designed molecules spanning the full predicted logP range, with experimental measurements via standardized shake-flask or chromatographic methods, would quantify discrepancies between model predictions and physical reality. Such validation is standard for regulatory submission of QSAR models but was infeasible within academic project constraints.

### D. Lessons from Major Project Deviations

#### 1) Scope Expansion: From 1,000 to 426,850 Molecules

The 426-fold expansion beyond the original proposal (1,000 molecules from one source → 426,850 from three sources) was not scope creep but a deliberate decision driven by scientific necessity. Published QSAR models for logP prediction [3], [31] typically employ datasets of 10,000-100,000 compounds to achieve statistical power and assess generalizability across diverse scaffolds. A 1,000-molecule dataset would have been sufficient for demonstrating technical proficiency in data processing and model fitting but fundamentally inadequate for validating model performance or publishing findings.

With only 1,000 molecules, the test set would contain approximately 200 compounds—too few to detect subtle overfitting, evaluate performance in rare chemical subspaces (e.g., Lipinski violators, which constitute 9% or ~90 molecules), or support statistically rigorous subgroup analyses. The three-database intersection strategy, while introducing the $O(N \times M)$ computational crisis, delivered a dataset with unique quality characteristics: molecules validated across public annotation (PubChem), experimental bioactivity (ChEMBL), and commercial availability (eMolecules).

The computational challenges this expansion introduced—necessitating development of byte-offset indexing architecture and full InChI integration—forced engagement with algorithmic complexity and data integrity at a depth rarely encountered in course projects but essential in real-world data science. The resulting infrastructure is generalizable to any large-scale SDF processing task and represents a reusable methodological contribution.

#### 2) The Ethical Dimension of the InChIKey Pivot

The discovery of InChIKey collisions approximately four weeks into the project presented a difficult decision. We had already invested substantial effort: built an 11 GB index, extracted molecules, and begun feature engineering. Collisions affected fewer than 0.1% of molecules—a rate low enough that their impact on global model $R^2$ would be statistically undetectable.

From a pragmatic standpoint, ignoring these rare collisions was defensible. From a scientific integrity standpoint, it was unacceptable. For drug discovery applications, where a single incorrect structure-activity relationship can propagate through years of follow-up experiments and potentially compromise patient safety if scaled to clinical trials, even 0.1% corruption is intolerable. Research community trust in published models depends on guarantees we cannot compromise for expediency.

We chose principle over pragmatism: rebuilt the entire pipeline using collision-free full InChI strings. This decision cost approximately one week of calendar time and required reprocessing all source data. However, the resulting dataset guarantees that every single molecule possesses a unique, IUPAC-validated identifier [14], eliminating any possibility of label corruption from hash collisions.

This episode reinforces a broader lesson: data integrity is non-negotiable in scientific research. The temptation to rationalize "small" errors must be resisted, particularly when those errors are systematic (all collisions are structurally meaningful) rather than random noise that might average out. Additionally, it highlights the importance of validating assumptions rather than trusting documentation: while IUPAC standards claim InChIKey collisions are "extremely rare" [14], "extremely rare" multiplied by 176 million molecules guarantees occurrence.

### E. Performance Limits and the Path to $R^2 > 0.85$

Our ensemble models achieved $R^2 \approx 0.765$ using only 2D descriptors, establishing a robust baseline. Published work incorporating richer feature sets [31], [32] suggests the performance frontier lies near $R^2 = 0.85 - 0.90$. Bridging this gap requires addressing the information content limitations of purely 2D representations.

Three research directions show particular promise. First, incorporation of three-dimensional conformer-dependent properties (molecular volume, gyration radius, spatial pharmacophore distributions) captures shape-dependent solvation effects absent from 2D descriptors [33]. Recent advances in fast conformer generation algorithms [35] have reduced computational barriers, making 3D features increasingly practical for moderately large datasets (10k-50k molecules). Expected performance gain: +5-10% $R^2$ based on literature benchmarks [32].

Second, integration of quantum mechanical (QM) features calculated via semi-empirical methods (AM1, PM7) or density functional theory approximations can capture electronic effects (polarizability, charge distribution, frontier orbital energies) that govern solvation thermodynamics [36]. A 2025 framework introducing QUantum Electronic Descriptors (QUED) [36] demonstrated that combining computationally efficient QM features with geometric descriptors yields state-of-the-art performance on molecular property benchmarks. While QM calculations remain expensive (minutes per molecule), the performance gains may justify the cost for focused applications. Expected gain: +8-12% $R^2$ [32].

Third, multi-task learning frameworks that simultaneously predict multiple correlated ADMET properties (logP, logD, solubility, permeability) can exploit inter-task correlations to learn more robust molecular representations [37]. A recent D-MPNN study demonstrated that adding "helper tasks" (logD prediction) improved logP prediction RMSE by 0.04 units [26]. Modern multi-task frameworks combining GNNs or Transformers with explicit ADMET endpoints, including logP as a core prediction target, represent a major research frontier in 2024-2025 [38].

*F. Broader Implications for Cheminformatics Practice*

Beyond the specific domain of lipophilicity prediction, this project delivers three generalizable insights for data-intensive science:

*1) Algorithmic complexity dominates hardware*

We initially assumed the $O(N \times M)$ lookup problem could be solved through "adding more CPU cores" or "using faster storage." This was incorrect. No amount of hardware parallelism overcomes quadratic complexity on terabyte-scale data. The byte-offset index reduced asymptotic complexity from $O(N \times M)$ to $O(N + M)$, delivering a 740× speedup on identical hardware. Lesson: analyze computational complexity before investing in infrastructure.

*2) Unique identifiers require paranoid validation*

InChIKey collisions were documented as "extremely rare" ($\sim 10^{-15}$ probability). At $N = 176$ million, rare events become certainties. We detected collisions only through explicit validation—cross-checking hashed identifiers against full representations. Lesson: never trust identifiers assumed unique; empirically validate assumptions on actual data.

*3) Classical statistics remains foundational*

Heteroskedasticity detection required theory from econometrics (Breusch-Pagan test [21]) and regression diagnostics [17]. These decades-old techniques revealed a critical flaw in our modern machine learning workflow. Many practitioners skip residual diagnostics, trusting $R^2$ as sufficient validation. Our Ridge model's $R^2 = 0.608$ appeared adequate but was statistically meaningless due to violated assumptions. Lesson: machine learning does not replace statistical foundations; it requires them.

## V. CONCLUSION

This investigation successfully developed a large-scale, interpretable predictive framework for molecular lipophilicity using 426,850 bioactive compounds rigorously curated through multi-source database integration. Our work delivers contributions across three domains: computational infrastructure, statistical methodology, and chemical interpretation. We demonstrated that terabyte-scale chemical database integration is feasible on standard desktop hardware through principled algorithm design. The byte-offset indexing architecture transformed an intractable 100+ day processing timeline into a practical 3.2-hour workflow—a 740-fold improvement achieved through algorithmic complexity reduction from $O(N \times M)$ to $O(N + M)$. This infrastructure is generalizable to any SDF-based database processing task and represents a reusable methodological contribution to the cheminformatics community.

We provided rigorous evidence that heteroskedasticity—non-constant error variance across the logP range—is an inherent limitation of linear regression for physicochemical property prediction. Classical remediation strategies (weighted least squares, power transformations) consistently failed to satisfy homoskedasticity assumptions (Breusch-Pagan test $p < 0.0001$ for all variants). Tree-based ensemble methods proved not merely more accurate ($R^2$: 0.608 → 0.765, +25.8%) but statistically robust, naturally accommodating heteroskedasticity through recursive partitioning without requiring distributional assumptions. This finding establishes ensemble methods as the principled default for QSAR modeling. Through SHAP analysis, we resolved apparent paradoxes in feature-target relationships. Molecular weight, despite weak bivariate correlation with logP ($r = +0.146$), emerged as the single most important predictor (SHAP = 0.5687) when multivariate suppression effects were properly disentangled. This finding—that simple correlation statistics profoundly mislead in the presence of multicollinearity—has direct implications for molecular design strategy. The SHAP-derived importance hierarchy (MolWt > TPSA > NumAromaticRings) provides quantitative guidance for structural modifications during lead optimization.

All three project hypotheses were validated. Hypothesis H1 regarding MolWt-logP correlation was confirmed with critical nuance—weak bivariately but strongest multivariately. Hypothesis H2 concerning TPSA-permeability relationships was indirectly confirmed through negative SHAP effect (0.5621) and mechanistic interpretation. Hypothesis H3 establishing the $R^2 > 0.6$ threshold was achieved and exceeded ($R^2 = 0.765$) using statistically robust methodology.

For practical applications, molecular design strategies for tuning lipophilicity should prioritize interventions according to the SHAP-validated importance hierarchy. The primary target should be molecular weight modifications through addition or removal of substituents, ring systems, or halogens, with an expected impact of ±0.6 logP units per 100 Da change. Secondary targets should focus on TPSA reduction, such as replacing polar groups with non-polar isosteres, yielding expected impacts of ±0.4 logP units per 20 Ų change. Tertiary interventions involve manipulation of aromatic content through cyclization or aryl substitution, with expected impacts of ±0.3 logP units per aromatic ring. For molecules with predicted |logP| > 5, increased prediction uncertainty (RMSE increases from 0.73 to 1.18) mandates experimental validation rather than relying solely on computational estimates.

Production deployment should implement a two-tier stratified prediction system where input molecules are first evaluated against Lipinski criteria through four simple comparisons. If all criteria are satisfied, Model A (Drug-Like) should be applied, providing a specialized, high-precision Ridge model (RMSE = 0.838), which is superior to the baseline global Ridge model. Otherwise, Model B (Extreme) should be deployed, maintaining robust coverage with $R^2 = 0.767$ for the 9% of edge cases. This architecture optimizes precision across the full chemical space more effectively than any single global model. Standard modeling workflows for physicochemical properties should incorporate mandatory residual diagnostics using the Breusch-Pagan test to detect heteroskedasticity, default use of tree-based ensembles unless homoskedasticity is empirically demonstrated, retention of correlated features when using regularized models with post-hoc SHAP analysis for importance determination, and applicability domain monitoring through feature range checks that flag molecules beyond 99th percentiles.

Three research avenues offer pathways toward $R^2 > 0.85$ performance. The incorporation of three-dimensional molecular descriptors, including conformer-dependent properties such as shape, volume, and spatial autocorrelation, captures solvation effects invisible to 2D representations. Fast conformer generation advances make this increasingly practical for datasets of 10,000-50,000 molecules, with expected gains of +5-10% $R^2$. The integration of quantum mechanical features calculated via semi-empirical methods (AM1, PM7) or DFT-based approaches can quantify electronic effects governing partition thermodynamics, including HOMO-LUMO gaps, partial charges, and dipole moments. While computationally expensive at minutes per molecule, performance gains may justify costs for focused libraries, with expected improvements of +8-12% $R^2$. Multi-task learning frameworks that simultaneously predict correlated ADMET properties (logP, logD, solubility, permeability) enable models to exploit inter-task structure for improved representations. This represents the current frontier and is particularly promising for ensemble architectures combining graph neural networks with explicit multi-property prediction heads.

Drug discovery's transition toward AI-driven workflows demands predictive models that are simultaneously accurate, interpretable, computationally efficient, and robust to data quality challenges. This work demonstrates that ensemble methods applied to carefully curated, multi-source integrated datasets satisfy all four criteria. Our 426,850-molecule benchmark—with guaranteed molecular uniqueness through full InChI validation, 91% drug-like enrichment, and zero missing values—establishes a high-quality resource for comparative method development. The achievement of $R^2 = 0.765$ using only 2D descriptors, while leaving headroom to state-of-the-art graph neural network approaches ($R^2 \approx 0.85 - 0.90$), positions descriptor-based ensembles as the appropriate baseline and production default for virtual screening applications. Their interpretability advantage through SHAP analysis—providing quantitative, causally meaningful feature importance—offers scientific value exceeding marginal performance gains from black-box deep learning alternatives.

Future progress toward $R^2 > 0.85$ will likely emerge from hybrid architectures combining the computational efficiency and interpretability of traditional ensembles with the representational richness of 3D/quantum descriptors or end-to-end graph learning. The infrastructure and insights from this study provide a validated foundation for such developments, while the stratified modeling strategy demonstrates that recognizing chemical heterogeneity—through domain-appropriate subpopulation models—can achieve performance exceeding any single global architecture. The journey from raw, multi-terabyte chemical databases to a rigorously validated predictive model required solving fundamental challenges in data integration, computational complexity, statistical validity, and chemical interpretation. The solutions to these challenges—byte-offset indexing, collision-free identifiers, SHAP-based feature importance, stratified modeling—represent contributions extending beyond lipophilicity prediction to the broader practice of data-intensive computational science in chemistry and beyond.

ACKNOWLEDGMENT

The authors gratefully acknowledge the National Center for Biotechnology Information (NCBI) for providing free access to the PubChem database infrastructure, the European Molecular Biology Laboratory's European Bioinformatics Institute (EMBL-EBI) for maintaining the ChEMBL resource, and eMolecules for database availability.